\documentclass[fleqn,10pt]{wlscirep}
\usepackage[utf8]{inputenc}
\usepackage[T1]{fontenc}
\usepackage{color}
\usepackage{tabularray}
\usepackage{caption}
\usepackage{subcaption}
\usepackage{graphicx}
\usepackage{xcolor}
\usepackage{booktabs} 
\usepackage{siunitx}
\usepackage{ragged2e}
\usepackage{tabularx}
\usepackage{adjustbox}
\usepackage{authblk}
\usepackage{capt-of}
\usepackage{booktabs}
\usepackage{footmisc}

\newcommand{\corrauth}{\textsuperscript{*}}
\newcommand{\equalcontrib}{\textsuperscript{$+$}}

\definecolor{Shark}{rgb}{0.121,0.137,0.156}
\definecolor{Jewel}{rgb}{0.094,0.501,0.219}

\title{Privacy-Preserving Federated Fraud Detection in Payment Transactions with NVIDIA FLARE}

\author[1]{Holger R. Roth\corrauth}
\author[2]{Sarthak Tickoo}
\author[5]{Mayank Kumar}
\author[1]{Isaac Yang}
\author[1]{Andrew Liu}
\author[2]{Amit Varshney}
\author[2]{Sayani Kundu}
\author[4]{Iustina Vintila}
\author[4]{Peter Madsgaard}
\author[4]{Juraj Milcak}
\author[1]{Chester Chen\corrauth}
\author[1]{Yan Cheng}
\author[1]{Andrew Feng}
\author[1]{Jeff Savio\corrauth}
\author[3]{Vikram Singh}
\author[5]{Craig Stancill}
\author[2]{Gloria Wan}
\author[5]{Evan Powell}

\author[4]{Anwar Ul Haq\equalcontrib\corrauth}
\author[2]{Sudhir Upadhyay\equalcontrib\corrauth}
\author[3]{Jisoo Lee\equalcontrib\corrauth}

\affil[1]{NVIDIA, Santa Clara, USA}
\affil[2]{J.P. Morgan, New York, USA}
\affil[3]{Bank of New York, New York, USA}
\affil[4]{Royal Bank of Canada, Toronto, Canada}
\affil[5]{DeepTempo, San Francisco, USA}

\begin{abstract}
Fraud-related financial losses continue to rise, while regulatory, privacy, and data-sovereignty constraints increasingly limit the feasibility of centralized fraud detection systems. Federated Learning (FL) has emerged as a promising paradigm for enabling collaborative model training across institutions without sharing raw transaction data. Yet, its practical effectiveness under realistic, non-IID financial data distributions remains insufficiently validated.

In this work, we present a multi-institution, industry-oriented proof-of-concept study evaluating federated anomaly detection for payment transactions using the NVIDIA FLARE framework. We simulate a realistic federation of heterogeneous financial institutions, each observing distinct fraud typologies and operating under strict data isolation. Using a deep neural network trained via federated averaging (FedAvg), we demonstrate that federated models achieve a mean F1-score of 0.903—substantially outperforming locally trained models (0.643) and closely approaching centralized training performance (0.925), while preserving full data sovereignty.

We further analyze convergence behavior, showing that strong performance is achieved within 10 federated communication rounds, highlighting the operational viability of FL in latency- and cost-sensitive financial environments. To support deployment in regulated settings, we evaluate model interpretability using Shapley-based feature attribution and confirm that federated models rely on semantically coherent, domain-relevant decision signals. Finally, we incorporate sample-level differential privacy via DP-SGD and demonstrate favorable privacy–utility trade-offs, achieving effective privacy budgets below $\epsilon = 10.0$ with moderate degradation in fraud detection performance. Collectively, these results provide empirical evidence that FL can enable effective cross-institution fraud detection, delivering near-centralized performance while maintaining strict data isolation and supporting formal privacy guarantees.
\end{abstract}

\begin{document}

\flushbottom
\maketitle
%
%

\par\noindent\corrauth\,Corresponding authors: \{hroth,chesterc,jsavio\}@nvidia.com; anwar.ulhaq@rbc.com; sudhir.x.upadhyay@jpmorgan.com; Jisoo@bny.com.\par
\par\noindent\equalcontrib\,These authors contributed equally.\par

\thispagestyle{empty}

\section{Introduction}
\label{sec:intro}

Fraud detection remains a central challenge in the Financial Services Industry (FSI), with recent analyses indicating that U.S. merchants incur an average cost of \$4.61 for every \$1 of fraudulent activity in the US~\cite{lexisnexis2025tcofEcom}. As digital transaction volumes continue to rise and fraud schemes become increasingly sophisticated, financial institutions face mounting pressure to deploy adaptive, data-driven detection systems~\cite{javelin2025idfraud,dal2017credit}. Yet conventional approaches rely heavily on centralized data aggregation, a strategy that is increasingly untenable under modern regulatory constraints, stringent data sovereignty requirements, and institutional risk postures~\cite{eu2016gdpr}. These limitations have historically hindered cross-organizational collaboration, resulting in fragmented detection capabilities and reduced effectiveness in identifying emerging fraud typologies.

\begin{figure}[htbp]
    \centering
    \includegraphics[width=0.5\linewidth]{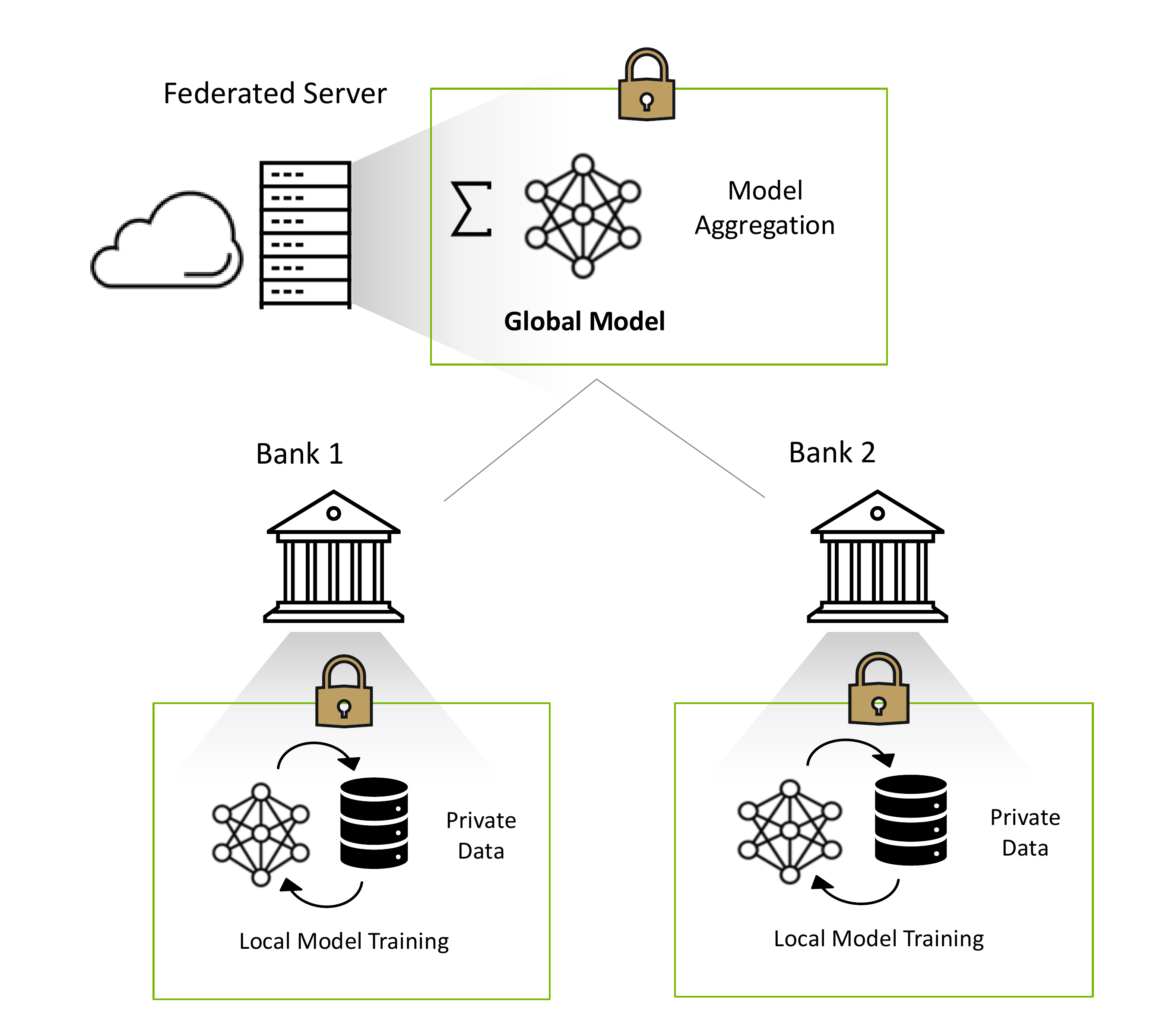}
    \caption{Federated learning for collaborative fraud detection.}
    \label{fig:fl_overview}
\end{figure}

Federated Learning (FL) provides a compelling alternative by enabling institutions to train shared machine learning models without exchanging raw customer or transaction data~\cite{li2020federated,kairouz2021advances}, as illustrated in Fig.~\ref{fig:fl_overview}. 
In principle, FL offers a mechanism for collective intelligence across institutions while preserving strict privacy guarantees and institutional autonomy. Yet, the practical feasibility of FL for fraud detection—characterized by highly imbalanced, non-IID, and institution-specific data distributions—has remained an open question~\cite{el2024federated,jpmorgan2025aikya}.

In this study, we present a comprehensive multi-party technical evaluation of FL for payment anomaly detection, implemented using the NVIDIA FLARE~\cite{Roth_NVIDIA_FLARE_Federated_2023} (NVFlare) federated computing framework. Our proof-of-concept experiment simulates a realistic multi-institution environment with heterogeneous fraud profiles, demonstrating that deep neural networks trained via federated averaging~\cite{mcmahan2017communication} (FedAvg) can achieve performance close to a centralized model (\textbf{0.903 vs 0.925}, $\Delta$F1=0.022), while improving substantially over local-only training (\textbf{0.643 $\rightarrow$ 0.903}, +40\%). Notably, this performance was accompanied by rapid convergence--typically within a few federated rounds--and strong cross-domain generalization across clients with disparate anomaly types.

To ensure trustworthiness and applicability in regulated financial contexts, we further validated the learned models using Shapley-value–based feature attribution explainability analysis~\cite{shapley1953value,lundberg2017unified}. Feature attribution studies confirmed that the federated model preserved interpretable, semantically coherent decision pathways, aligning with domain expectations and regulatory requirements for transparent AI~\cite{lundberg2017unified}.

Beyond demonstrating accuracy and scalability, we also evaluated privacy-enhancing mechanisms through \textbf{differentially private (DP) training}~\cite{abadi2016deep} and implemented secure sampling strategies to achieve sample-level differential privacy. Our findings indicate that private FL can achieve \textbf{privacy amplification by subsampling}~\cite{balle2018privacy}, significantly reducing effective privacy loss while maintaining competitive model performance. This result underscores the feasibility of integrating formal privacy guarantees into federated fraud detection pipelines, offering a practical path toward regulatory-grade privacy preservation.

Collectively, this work provides rigorous empirical evidence that FL represents a practical, high-performance, and privacy-preserving paradigm for multi-institution fraud detection. Our contributions include:

\begin{itemize}[noitemsep]
    \item Demonstration of collaborative deep learning across multiple simulated financial institutions without sharing sensitive data.
    \item Quantitative benchmarking of FedAvg against central \& local-only training, establishing strong performance baselines in non-IID settings.
    \item Validation of interpretability through Shapley-value-based attribution analysis to support transparency to foster regulatory compliance.
    \item Assessment of differential privacy techniques, showing promising privacy-utility trade-offs and measurable privacy amplification effects.
    \item Evaluation of scalability, communication efficiency, and operational readiness of NVFlare in a high-security, distributed FSI environment.
\end{itemize}

This study illustrates that FL can deliver near-centralized model performance while preserving absolute data sovereignty, enabling a new generation of collaborative fraud detection systems that align with both technological and regulatory imperatives.

\section{Results}
\label{sec:results}

We develop a purely synthetic data generation framework providing controllable statistical heterogeneity across federated sites, composable rule-based anomaly injection, and adjustable class balance with label noise (see \hyperref[sec:methods]{Methods}). Each institution received a dataset with a distinct distribution of fraud categories, creating a non-IID environment representative of real-world inter-institutional variation.

Each client operated on three dataset partitions:
\begin{enumerate}[noitemsep]
    \item \textbf{Scaling Dataset:} Used to compute feature normalization parameters applicable across all anomaly categories.
    \item \textbf{Training Dataset:} Used for local FL rounds.
    \item \textbf{Test Dataset:} Reserved for local model validation and performance monitoring.
\end{enumerate}

\begin{figure}[!htbp]
\centering

\begin{subfigure}[t]{0.48\textwidth}
    \centering
    \footnotesize
    \caption{Fraud rates across participating banks, with the highest fraud rate at Site-E. Average transaction amounts are shown in thousands of USD, while the total amounts are shown in billions of USD. \label{tab:fraud_rates}}
    \begin{tabular}{llllll}
    \toprule
    Site   & Total    & Fraud & Rate   & Avg        & Total      \\
           & Records  & Count & (\%)   & Amount     & Amount     \\
    \midrule
    Site-A & 200,000  & 1,077 & 0.54   & \$40.28K   & \$8.06B    \\
    Site-B & 200,000  & 1,060 & 0.53   & \$36.99K   & \$7.40B    \\
    Site-C & 200,000  & 1,162 & 0.58   & \$25.28K   & \$5.06B    \\
    Site-D & 200,000  & 1,196 & 0.60   & \$336.98K  & \$67.40B   \\
    Site-E & 200,000  & 995   & 0.50   & \$1768.23K & \$353.65B  \\
    \bottomrule
    \end{tabular}
\end{subfigure}
\hfill
\begin{subfigure}[t]{0.48\textwidth}
    \vspace{0pt} 
    \centering
    \includegraphics[width=\linewidth]{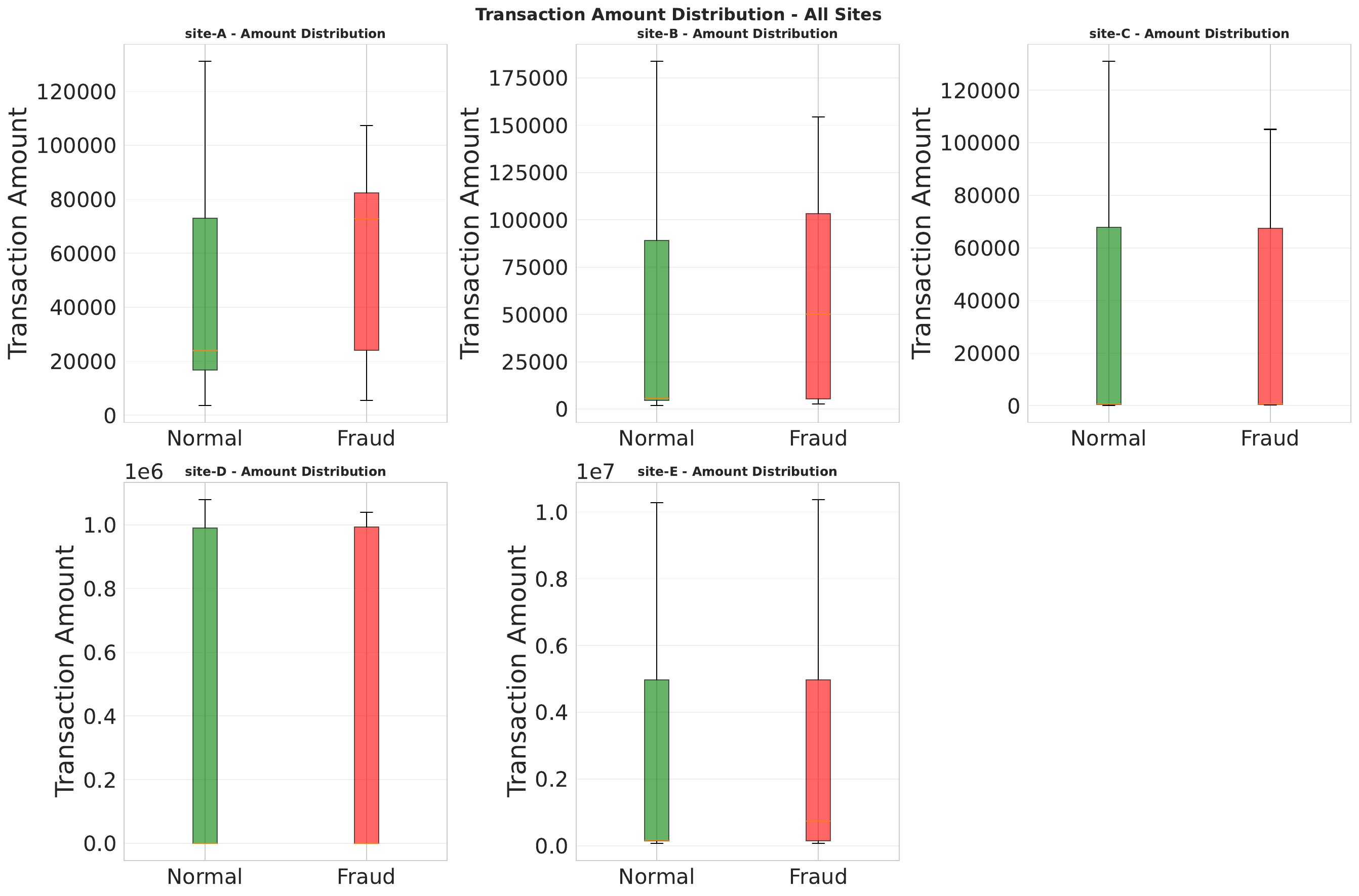}
    \caption{USD amounts for normal and fraudulent transactions.\label{fig:amounts}}
\end{subfigure}

\vspace{1em}

\begin{subfigure}[t]{0.48\textwidth}
    \centering
    \includegraphics[width=\linewidth]{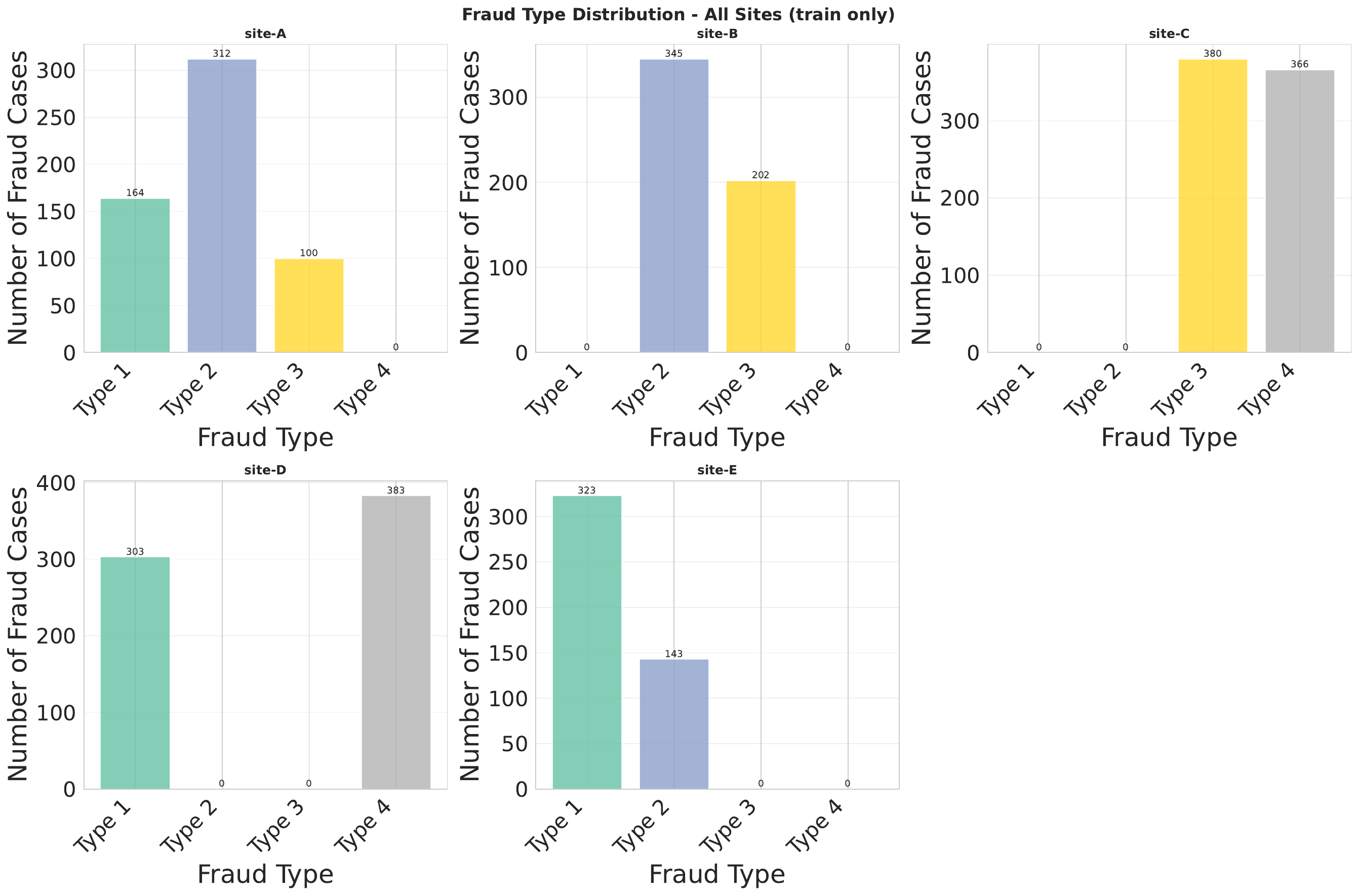}
    \caption{Fraud types (train).\label{fig:fraud_types_train}}
\end{subfigure}
\hfill
\begin{subfigure}[t]{0.48\textwidth}
    \centering
    \includegraphics[width=\linewidth]{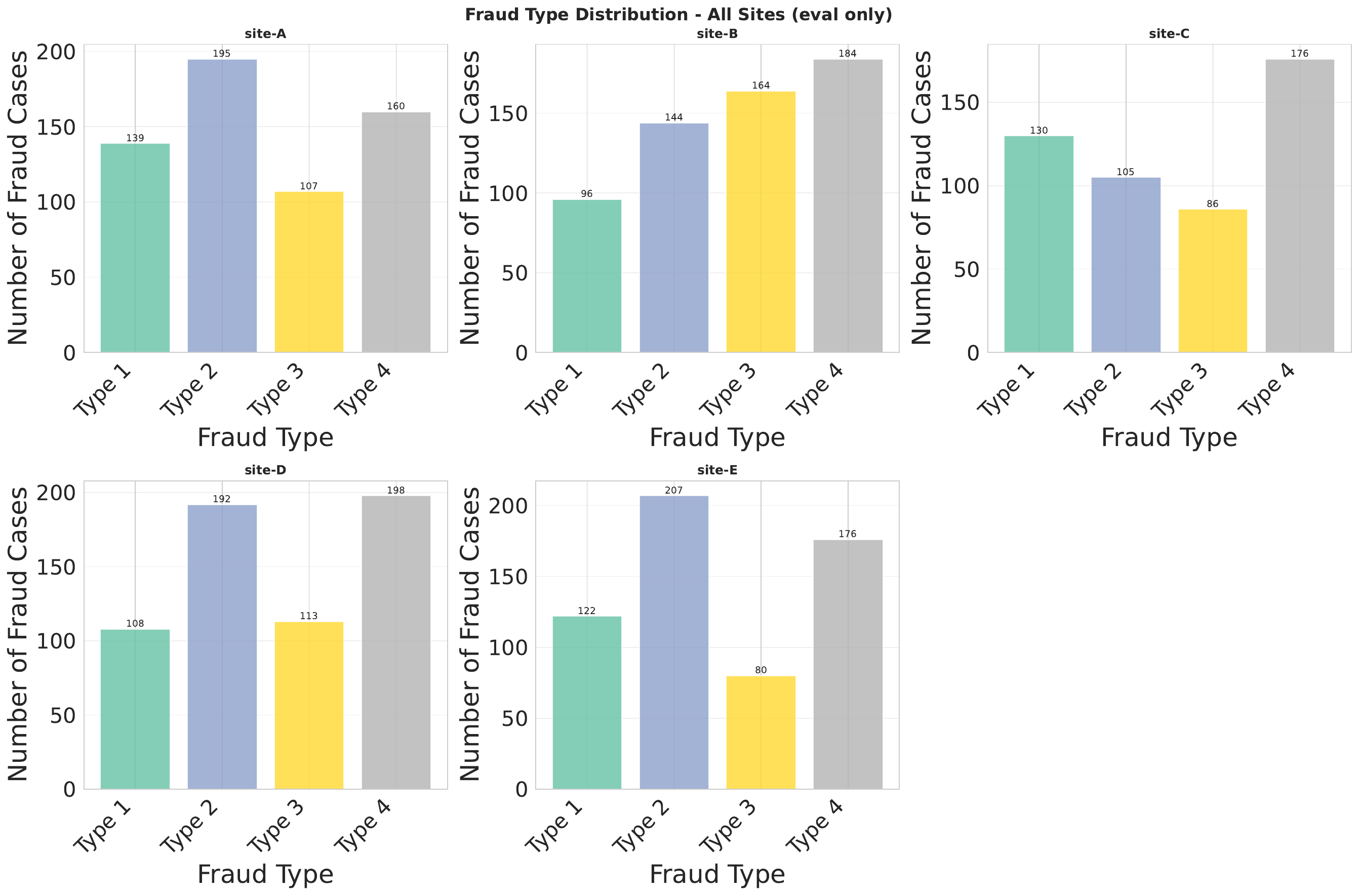}
    \caption{Fraud types (evaluation).\label{fig:fraud_types_eval}}
\end{subfigure}

\caption{\textbf{Data distributions across sites.}
\textbf{A,} Site-level fraud rates and aggregate amount statistics.
\textbf{B,} Amount distributions for normal vs fraudulent transactions.
\textbf{C--D,} Anomaly-type distributions for training and evaluation partitions.}
\label{fig:distributions}
\end{figure}

This configuration ensures controlled experimentation while faithfully representing the heterogeneity encountered in operational fraud detection systems. Figures~\ref{fig:fraud_types_train} and ~\ref{fig:fraud_types_eval} show the distribution of fraud types across the different participating banks for training and evaluation sets, respectively.
To further characterize the heterogeneity of the simulated federation, we examined the distributional properties of transaction amounts and the prevalence of fraud across institutions. Fig.~\ref{tab:fraud_rates} summarizes key dataset statistics—including total volume, fraud counts, and fraud rates—which together underscore the non-IID nature of the participating clients and the resulting challenges for federated optimization. Fraud rates are computed on the combined training+test from the final fraud labels after label-noise application.
Complementing this, Fig.~\ref{fig:amounts} illustrates the variation in USD payment amounts between fraud and non-fraud transactions at each site, highlighting substantial inter-institutional differences in both scale and dispersion.

\subsection{Data Configuration}

\clearpage
\subsection{Experiment 1: Federated Learning Performance and Convergence}

The federated model trained using FedAvg~\cite{mcmahan2017communication} consistently outperformed locally trained models across all clients and fraud types. The average F1-score across clients increased from 0.643 (local training) to 0.903 (FedAvg), representing an improvement of 40\% as shown in Table~\ref{tab:all_f1}. We furthermore compare advanced FL algorithms such as FedOpt~\cite{reddi2020adaptive}, which uses an SGD optimizer on the server to update the global model, and FedProx~\cite{li2020federated}, which adds a proximal term during local training to prevent models from drifting too far from the current global model. Our results indicate that these advanced algorithms are not necessary to achieve a higher performance in our scenario.

\begin{table}[!htbp]
\centering
\caption{Mean binary F1 after 20 rounds across all clients and fraud types.}
\label{tab:all_f1}
\small
\begin{tabular}{@{}l S[table-format=1.3]@{}}
\toprule
\textbf{Algorithm} & {\textbf{Mean F1-score}} \\
\midrule
Central & 0.925 \\
FedAvg  & 0.903 \\
FedOpt  & 0.905 \\
FedProx & 0.903 \\
Local   & 0.643 \\
\bottomrule
\end{tabular}
\end{table}

\begin{figure}[!htbp]
    \centering
    \begin{subfigure}[t]{0.55\linewidth}
        \centering
        \includegraphics[width=\linewidth]{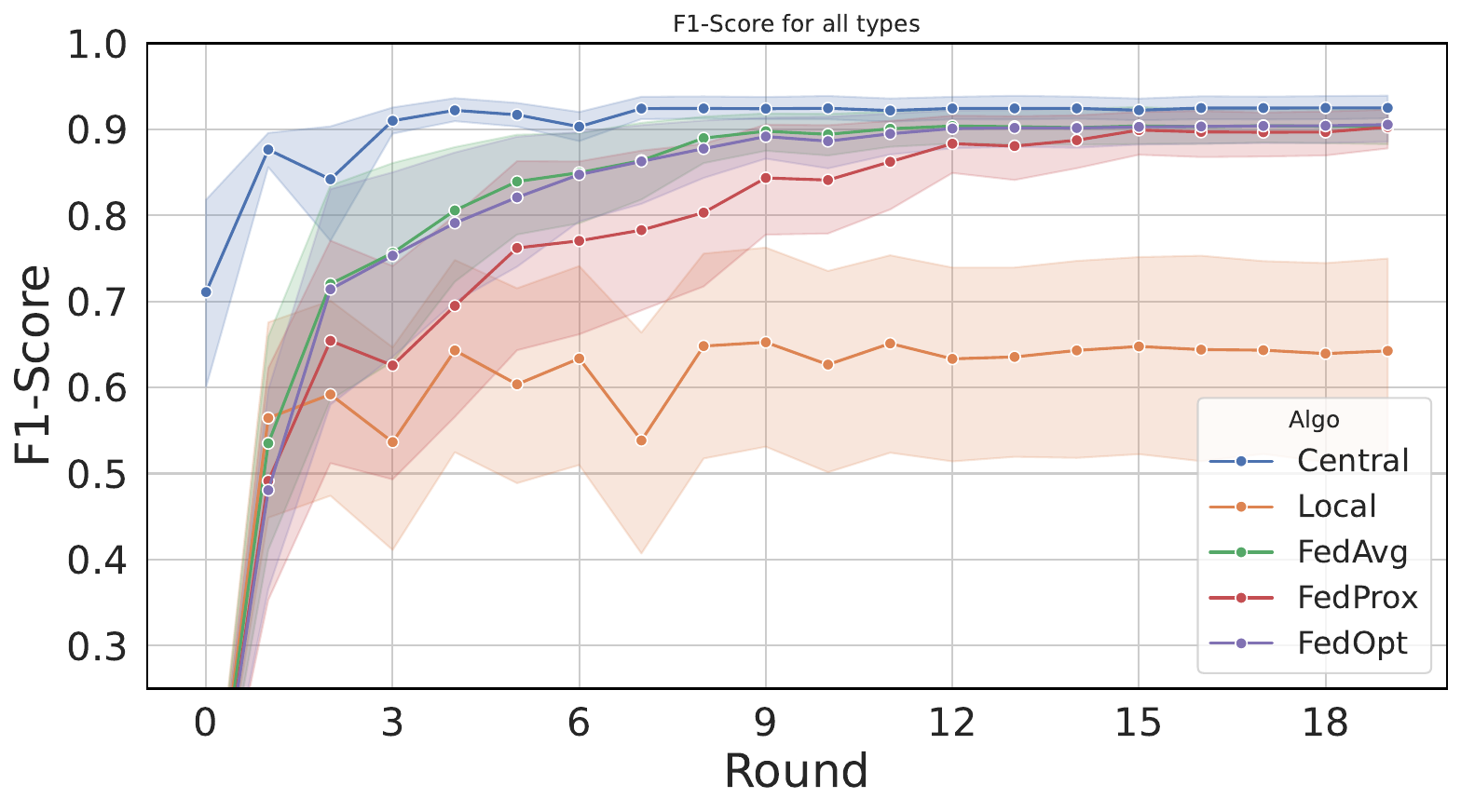}
        \caption{}
    \end{subfigure}\hfill
    \begin{subfigure}[t]{0.45\linewidth}
        \centering
        \includegraphics[width=\linewidth]{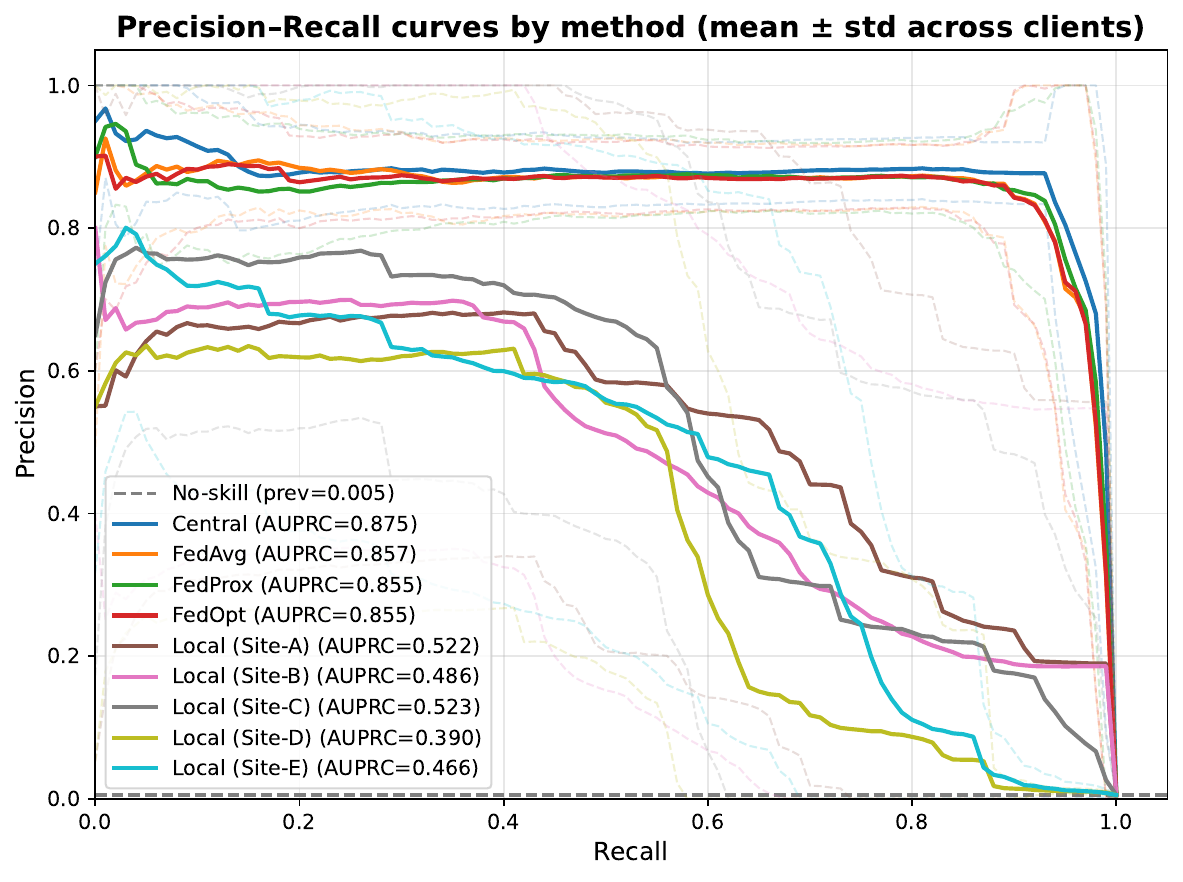}
        \caption{}
    \end{subfigure}
    \caption{\textbf{Central vs. Local vs. FL: convergence and precision--recall performance.}
    \textbf{A,} Mean binary F1-score across clients and anomaly types over federated rounds (shaded bands: s.d.\ across clients).
    \textbf{B,} Precision--recall curves evaluated across client test sets for each checkpointed model; solid lines show the mean and bands indicate $\pm$1 s.d.\ across clients. The no-skill baseline is included, and AUPRC values are reported in the legend.}
    \label{fig:all_f1}
\end{figure}

When comparing model convergence over training rounds, FedAvg maintains a consistent advantage over the more advanced federated optimizers, as shown in Fig.~\ref{fig:all_f1}A. In addition, the precision--recall curves (Fig.~\ref{fig:all_f1}B) confirm improved performance under class imbalance, with FedAvg approaching centralized performance and substantially outperforming locally trained models.

\begin{figure}[!htbp]
    \centering
    \begin{subfigure}[b]{0.45\textwidth}
        \centering
        \includegraphics[width=\linewidth]{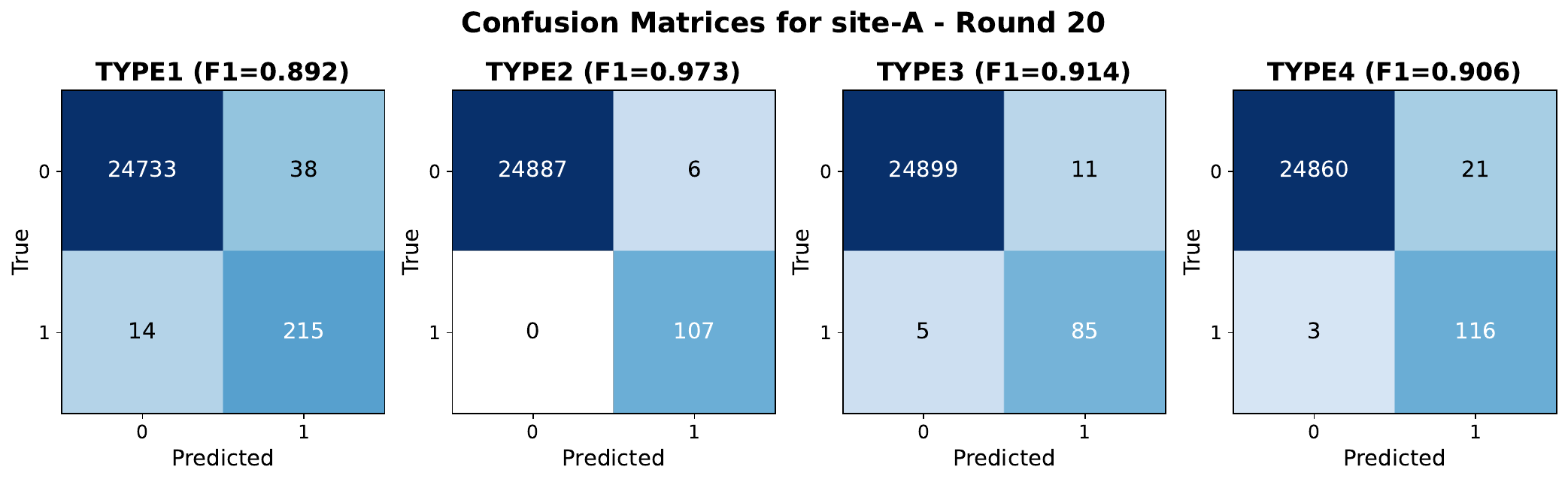} \\
        \caption{Site-A}
    \end{subfigure}\hfill    
    \begin{subfigure}[b]{0.45\textwidth}
        \label{fig:cm_site-A}
        \centering
        \includegraphics[width=\linewidth]{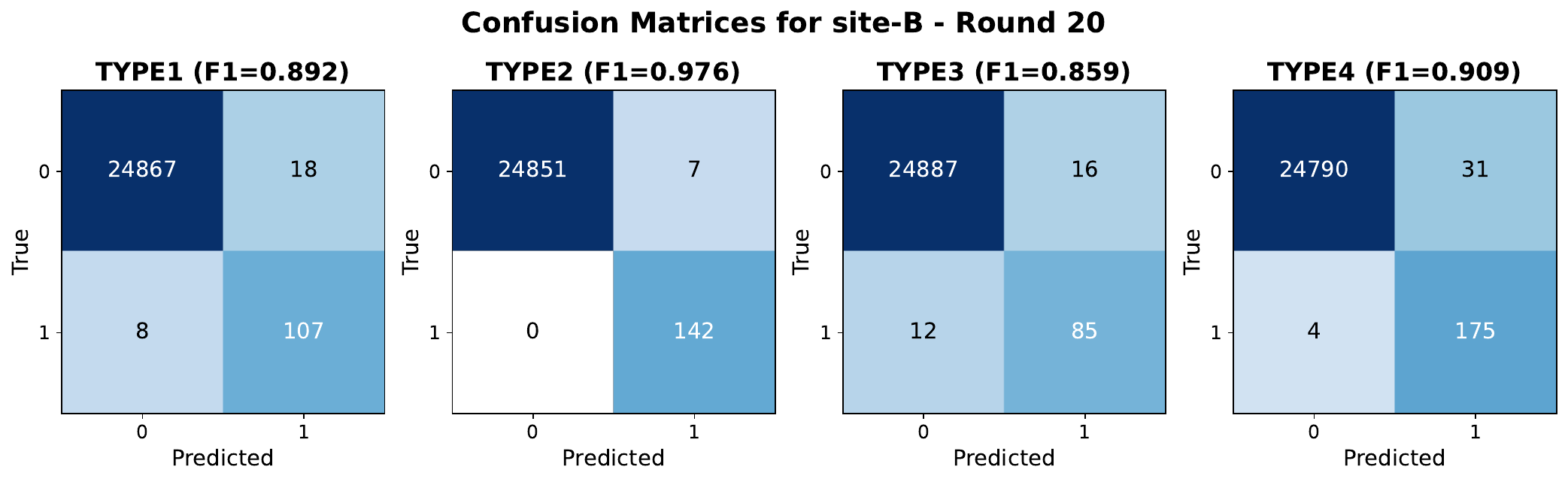} \\
        \caption{Site-B}
        \label{fig:cm_site-B}    
    \end{subfigure}        
    \begin{subfigure}[b]{0.45\textwidth}
        \centering
        \includegraphics[width=\linewidth]{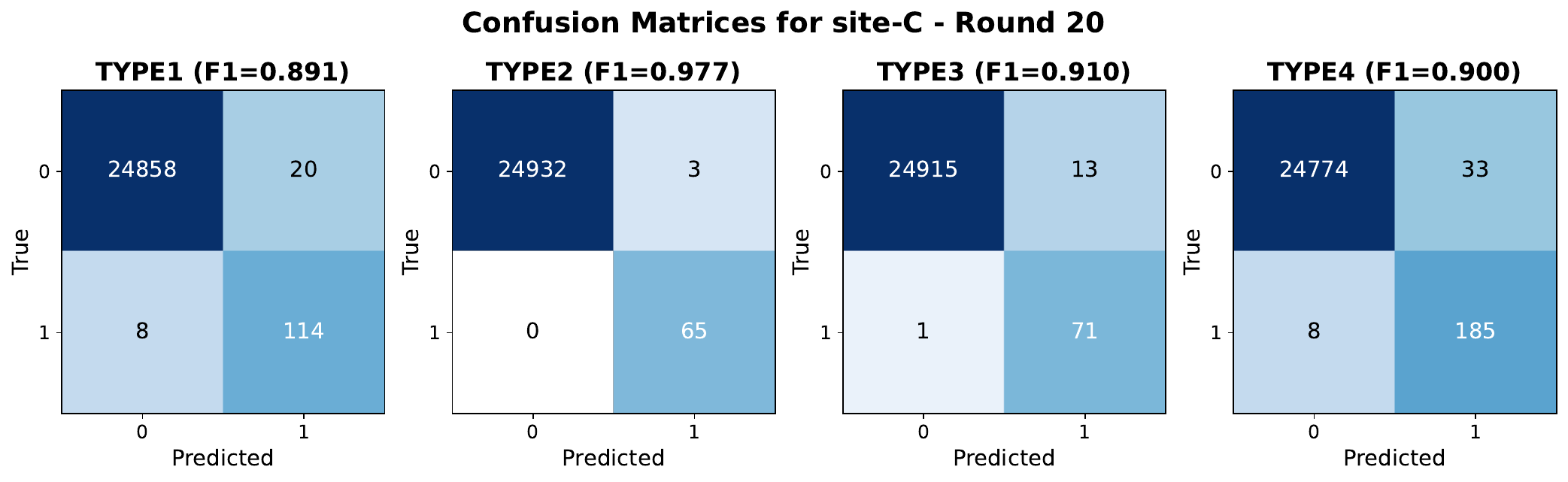} \\
        \caption{Site-C}
        \label{fig:cm_site-C}    
    \end{subfigure}\hfill        
    \begin{subfigure}[b]{0.45\textwidth}    
        \centering
        \includegraphics[width=\linewidth]{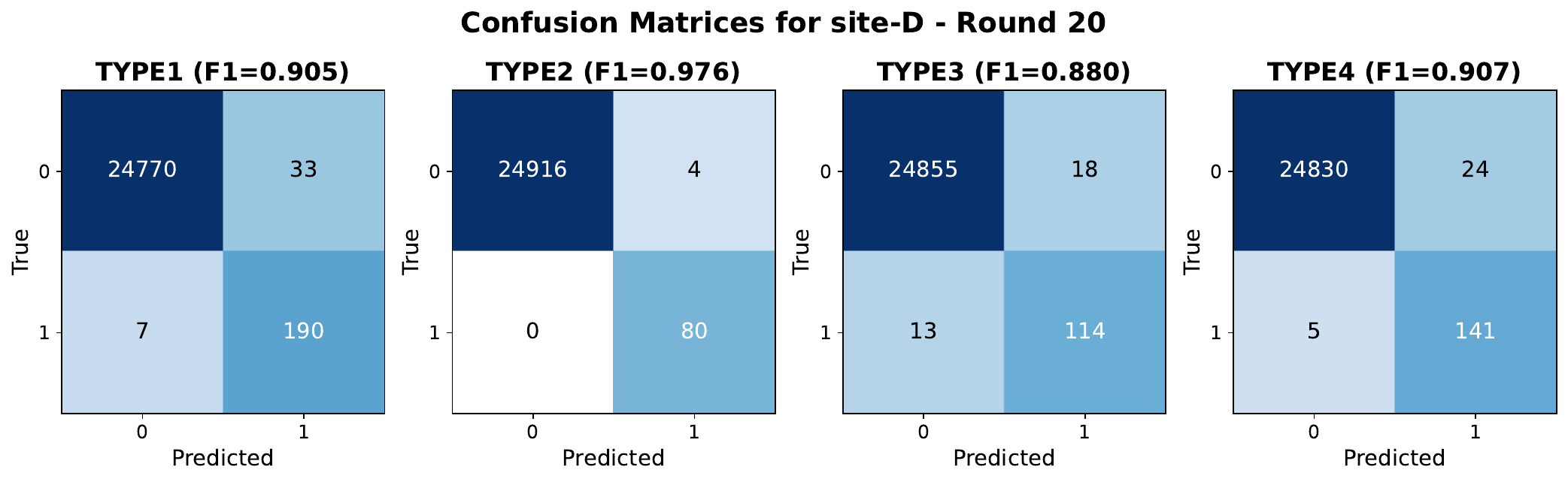} \\
        \caption{Site-D}
        \label{fig:cm_site-D}        
    \end{subfigure}        
    \begin{subfigure}[b]{0.45\textwidth}    
        \centering
        \includegraphics[width=\linewidth]{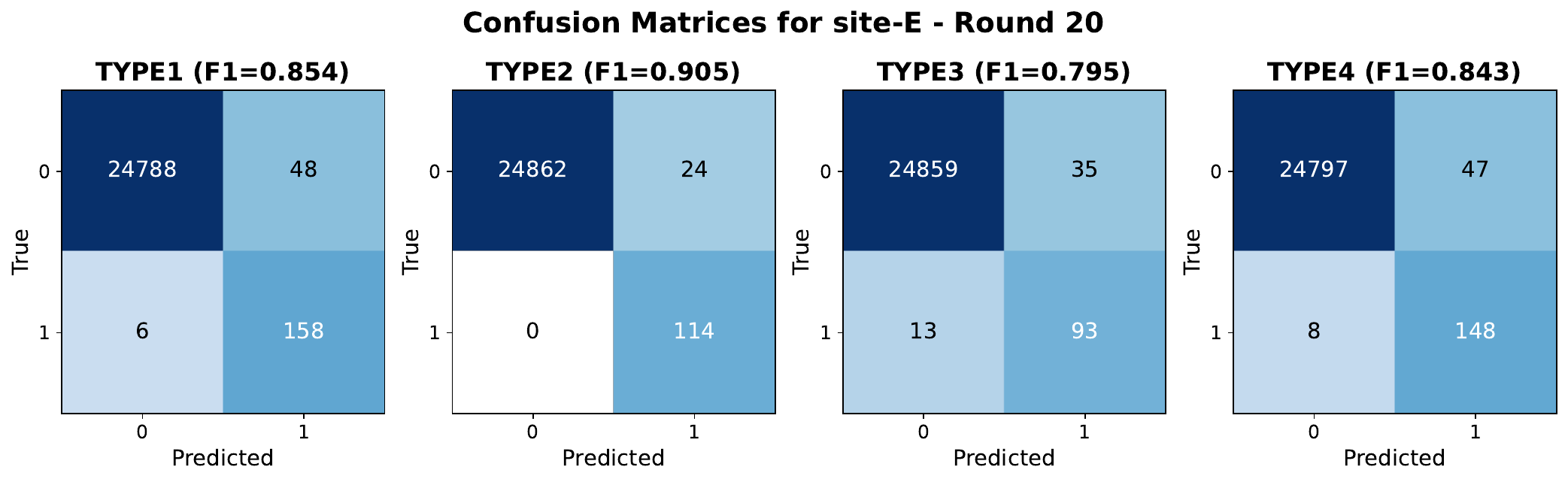} \\
        \caption{Site-E}
        \label{fig:cm_site-E}            
    \end{subfigure}
    \caption{Confusion matrices of the final FedAvg global model for each participating bank.}
    \label{fig:cm}
\end{figure}

Next, we examine the confusion matrices derived from the final global model trained with FedAvg for each of the five institutions (Fig.~\ref{fig:cm}). For this ‘per-type’ evaluation, we compute binary F1 on the subset corresponding to a given anomaly type vs. normal (one-vs-rest) while in Fig.~\ref{fig:all_f1}, we report the mean across types and clients. The matrices at round 20 reveal consistently strong true-positive rates and notably low false-positive rates across most fraud categories. These results indicate that the federated model effectively balances precision and recall, achieving high sensitivity to fraudulent activity while minimizing misclassifying legitimate transactions. Such a balance is essential for operational fraud-detection systems, where both failing to identify fraudulent behavior (false negatives) and erroneously blocking legitimate transactions (false positives) incur substantial financial and customer-experience costs.

In summary, we can observe the following takeaway from the quantitative results.
\begin{itemize}
    \item \textbf{Performance Superiority:} FedAvg achieved a higher performance than any of the banks training on their local data alone, while achieving a performance comparable to centralized training, validating its ability to improve fraud detection while preserving data privacy collaboratively.
    \item \textbf{Rapid Convergence:} Learning curves show models stabilize quickly, typically within 3-5 federated training rounds, indicating efficient aggregation of knowledge from heterogeneous data sources with the potential to reduce training times.
    \item \textbf{Cross-Domain Learning:} Clients demonstrated significant performance improvement on fraud types not present in their local training data. For instance, clients with primarily Type 1 fraud data achieved high F1-scores (>0.94) on Type 2 fraud after federated training.
\end{itemize}

\subsection{Model Interpretability and Feature Importance}
\begin{figure}[!htbp]
    \centering
    \begin{subfigure}[b]{0.45\textwidth}
        \centering
        \includegraphics[width=\linewidth]{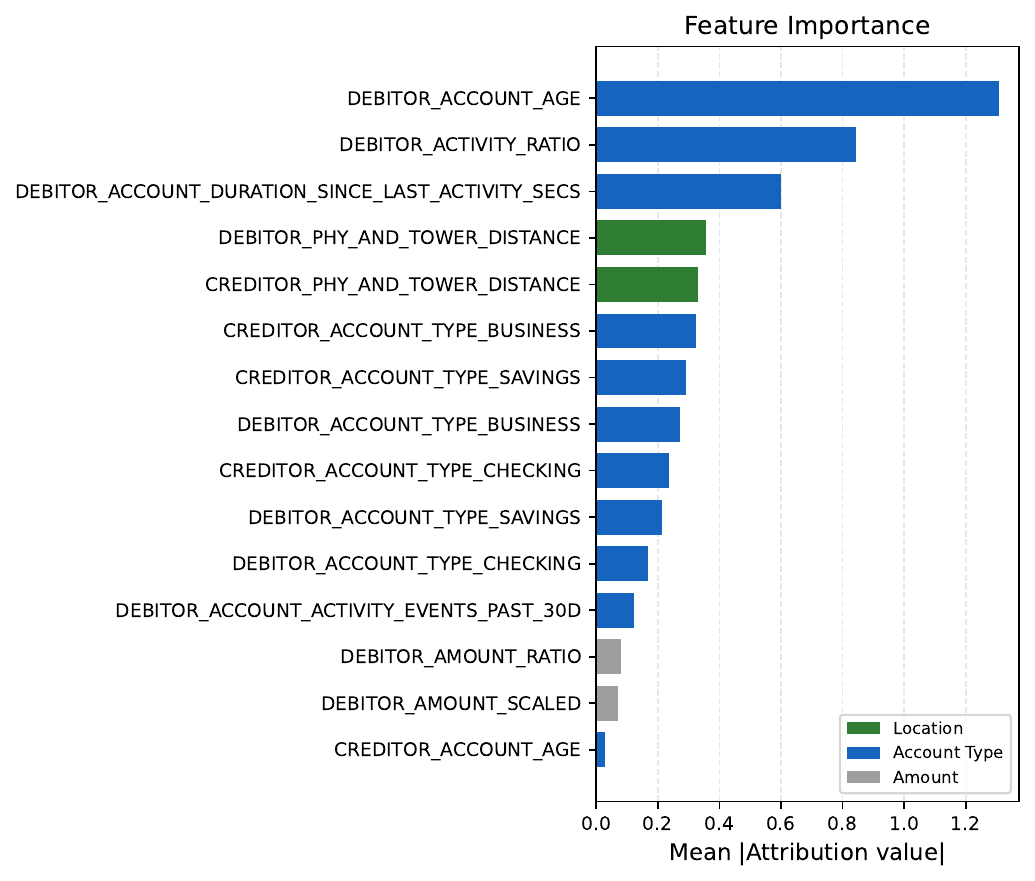} \\
        \caption{Central Training}
    \end{subfigure}
    \begin{subfigure}[b]{0.45\textwidth}
        \label{fig:feature_importance_central}
        \centering
        \includegraphics[width=\linewidth]{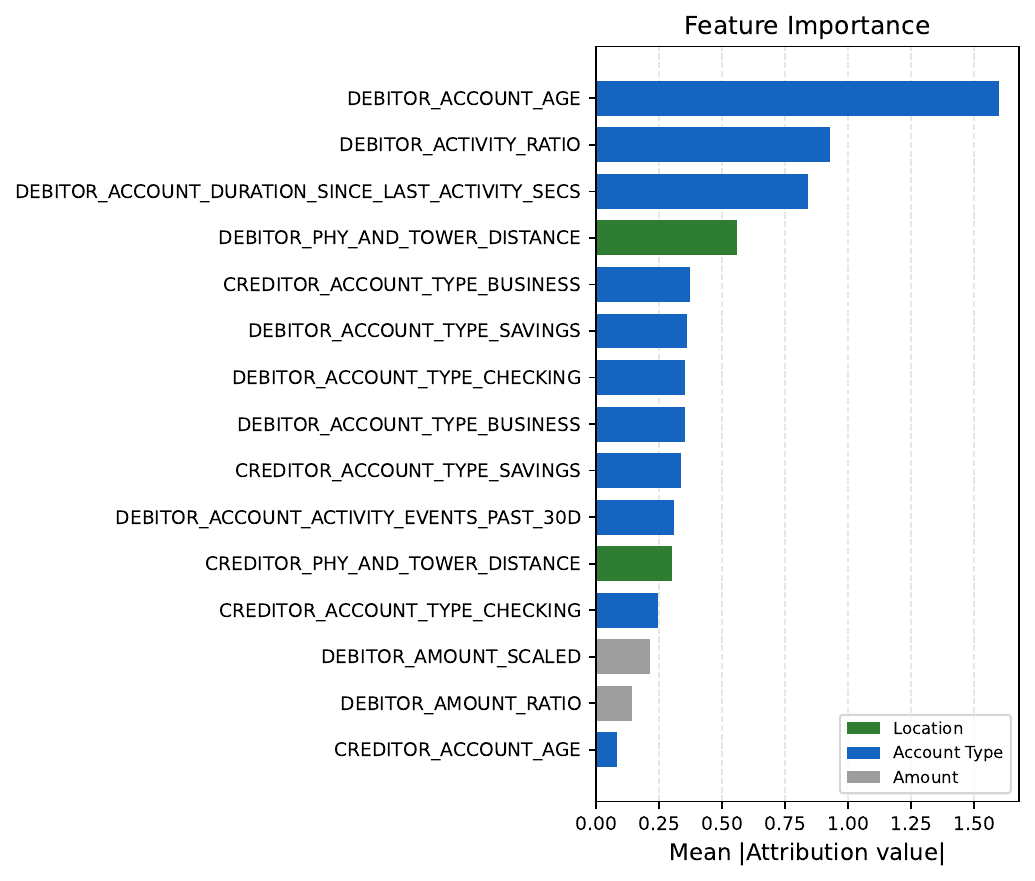} \\
        \caption{Federated Learning}
        \label{fig:feature_importance_fedavg}    
    \end{subfigure}
    \caption{Feature importance plot based on Shapley-value–based attribution (Captum GradientShap).}
    \label{fig:feature_importance}
\end{figure}

Feature importance analysis using Captum\footnote{\url{https://captum.ai}} confirmed that the federated model learned meaningful patterns that align with domain expectations, as illustrated in Figure~\ref{fig:feature_importance}.



For location-based fraud (Type 1), geographical features dominated the importance scores. For account age and transaction amount-based fraud (Type 2), those features were prioritized, confirming that the model’s decisions were interpretable and grounded in relevant financial indicators.

\clearpage

\subsection{Experiment 2: Federated Learning with Differential Privacy}

\begin{figure}[!htbp]
    \centering
    \includegraphics[width=0.5\linewidth]{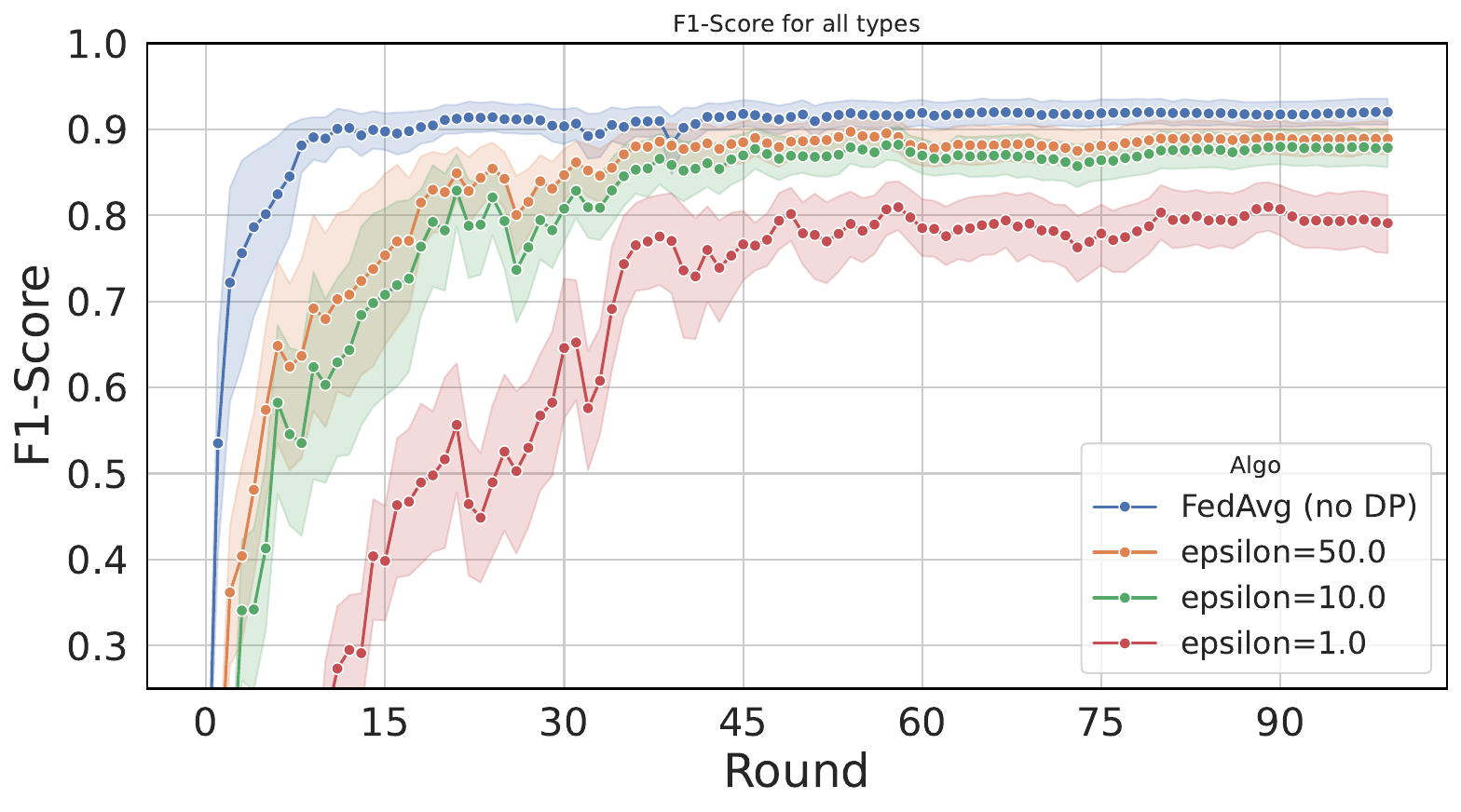}
    \caption{\textbf{FedAvg vs. DP-FedAvg:} F1-scores for all fraud types averaged across clients. The bands indicate the std. dev. of the metrics.}
    \label{fig:all_f1_dp}
\end{figure}

Incorporating differentially private stochastic gradient descent (DP-SGD) yielded strong formal privacy guarantees, achieving an effective privacy budget of $\epsilon = 10.0$, while maintaining a competitive model performance as shown in Fig.~\ref{fig:all_f1_dp}. These findings highlight that robust privacy preservation can be integrated into FL pipelines with moderate degradation in utility. We utilize the Opacus\footnote{\url{https://opacus.ai/}} library~\cite{yousefpour2021opacus} to implement DP-SGD on each client during local training. The privacy budget accumulates across FL rounds, and each epsilon value represents the exhausted budget for the entire FL training run. Final model performances are shown in Table~\ref{tab:all_f1_dp}.

\begin{table}[!htbp]
\centering
\caption{\textbf{Mean F1-score after Round 100 with DP-SGD.} Mean binary F1 across all clients and fraud types for different privacy budgets.}
\label{tab:all_f1_dp}
\small
\begin{tabular}{@{}l l S[table-format=1.3]@{}}
\toprule
\textbf{Algorithm} & \textbf{Privacy} & {\textbf{Mean F1-score}} \\
\midrule
FedAvg & no DP           & 0.920 \\
FedAvg & $\epsilon=50.0$ & 0.889 \\
FedAvg & $\epsilon=10.0$ & 0.879 \\
FedAvg & $\epsilon=1.0$  & 0.791 \\
\bottomrule
\end{tabular}
\end{table}

\clearpage

\section{Discussion}
\label{sec:discussion}

This study evaluated federated learning (FL) for collaborative fraud 
detection across heterogeneous financial institutions using NVIDIA FLARE, 
addressing four questions: FL performance relative to centralized training, 
convergence speed, model interpretability, and privacy-utility trade-offs 
under differential privacy.

\textbf{Performance and Cross-Institution Collaboration.} FedAvg achieved 
a mean F1-score of 0.903 after 20 rounds, closely approaching centralized 
performance (0.925, $\Delta$F1 = 0.022) while substantially outperforming 
local-only training (0.643, +40\%). The primary driver of this improvement 
is cross-domain generalization: each institution observed only a subset of 
fraud types during local training, mirroring real-world conditions where 
individual institutions encounter limited fraud typologies. Federation 
resolved this directly, with clients achieving F1 $>$ 0.94 on fraud types 
absent from their local training data, as confirmed by the per-type 
confusion matrices at round 20 (Fig.~\ref{fig:cm}). This result suggests 
that the primary operational value of FL in financial fraud detection is 
not incremental improvement on known fraud types, but rather the ability 
to detect emerging or underrepresented typologies that a single institution 
would otherwise miss.

\textbf{Choice of Methods.} DNNs were preferred over tree-based approaches 
such as gradient boosting because their gradient-based, continuous parameter 
spaces are directly compatible with FedAvg-style weight averaging---a 
property tree-based models do not share. The compact three-layer architecture 
(64, 32, 16 neurons) balances model expressiveness against per-round 
communication overhead. FocalLoss was selected over standard cross-entropy 
to address the severe class imbalance across all sites (fraud rates 
0.50--0.60\%, Fig~\ref{tab:fraud_rates}), focusing gradient updates on 
the harder fraud minority rather than the dominant legitimate transaction 
class.

Among FL aggregation algorithms, FedAvg, FedProx, and FedOpt converged to 
similar final F1-scores (0.903, 0.903, 0.905 respectively, 
Table~\ref{tab:all_f1}), with FedAvg showing the most stable trajectory. 
FedProx's proximal regularization and FedOpt's server-side momentum are 
most beneficial under severe client drift, which was not the dominant 
condition here given the structurally similar feature distributions across 
sites. FedAvg is therefore the appropriate default for federations with 
moderate heterogeneity; more complex algorithms should be considered when 
fraud typologies vary significantly, or data volumes are highly asymmetric 
across institutions.

\textbf{Convergence.} Models stabilized within 3--5 federated rounds, 
with marginal gains beyond round 10 (Fig.~\ref{fig:all_f1}A). This rapid 
convergence limits communication overhead to a short initial window, making 
FL operationally viable in latency- and cost-sensitive financial 
environments. The NVFlare Scatter-Gather workflow deployed on H200 GPU 
instances via NVIDIA DGX Cloud supported 
this cadence without observable bottlenecks in the experimental 
configuration.

\textbf{Interpretability and Regulatory Alignment.} GradientShap 
attribution confirmed that the federated model's decision signals align 
with the underlying fraud rules: location features dominated for Type 1 
fraud, while account age and transaction amount led for Type 2 
(Fig.~\ref{fig:feature_importance}). The close alignment between federated 
and centralized attribution profiles indicates that federated aggregation 
does not distort learned representations relative to a model trained on 
pooled data. These results support the explainability requirements of 
model risk management frameworks such as SR 11-7 within the scope of 
this proof-of-concept, where model decisions must be grounded in 
domain-consistent and auditable feature signals.

\textbf{Privacy-Utility Trade-offs.} DP-SGD introduced monotonic F1 
degradation as privacy budgets tightened: from 0.920 (no DP) to 0.879 
($\varepsilon = 10.0$) and 0.791 ($\varepsilon = 1.0$) over 100 rounds 
(Table~\ref{tab:all_f1_dp}). The practical acceptability of this 
trade-off depends on each institution's operating threshold---at high 
transaction volumes, even a 0.04 F1 drop has measurable impact on fraud 
caught and false alarms generated, and this should be evaluated against 
the institution's specific risk tolerance before deployment. It is also 
important to distinguish the two privacy mechanisms used in this study. 
DP-SGD provides sample-level protection against inference attacks on the 
trained model, while FL's architectural guarantee ensures raw transaction 
data never leaves each institution. These address distinct threat surfaces, 
and their combination provides a more complete privacy posture than either 
mechanism alone---a relevant consideration for institutions navigating 
compliance requirements under frameworks such as GDPR and data sovereignty 
regulations.

\textbf{Limitations.} Three constraints bound the scope of these results. 
First, all experiments used synthetic data with rule-based anomaly 
injection, which does not capture real-world fraud complexity including 
adversarial adaptation, concept drift, or ground truth ambiguity---validation 
on real transaction data remains a necessary step before production 
deployment. Second, the federation comprised five sites with equal data 
volumes and clean fraud-type partitions; real deployments will involve 
greater asymmetry and overlapping typologies. Third, honest participant 
behavior was assumed throughout; robustness to malicious gradient updates 
or model poisoning was not evaluated. Follow-on work should prioritize 
shadow-mode pilot validation on real transaction data, stress-testing under 
more severe non-IID conditions, and integration of secure aggregation 
protocols to address the honest-participant assumption.

\textbf{Conclusion.} Overall, the findings reported in this study show strong evidence that FL—augmented with differential privacy—as a viable and scalable solution for collaborative fraud detection, capable of achieving near-centralized performance while preserving data sovereignty and offering higher privacy guarantees.

\section{Methods}
\label{sec:methods}

\subsection{Experiment Objectives}

The experiment was designed with the following design goals:
\begin{enumerate}
    \item \textbf{Non-IID Data Distribution:} Simulate realistic, heterogeneous data distributions where institutions encounter distinct fraud patterns.
    \item \textbf{Cross-Domain Generalization:} Evaluate the FL model’s ability to detect fraud types not present in a participant’s local data.
    \item \textbf{Privacy-Preserving Collaboration:} Validate effective model training without sharing raw transaction data.
    \item \textbf{Scalability Assessment:} Test the framework with multiple participants (5 client sites) on NVIDIA DGX Cloud.
    \item \textbf{Algorithm Comparison:} Benchmark FedAvg, FedProx, and FedOpt to identify optimal aggregation strategies.
    \item \textbf{Model Interpretability:} Ensure explainable predictions via Shapley analysis.
\end{enumerate}

\subsection{Data}

\paragraph{Synthetic Transaction Generator.}
Synthetic datasets were generated to simulate realistic payment transaction scenarios and ensure privacy compliance, controlled experimentation, and reproducibility. The data generation methodology was designed to support the overarching experimental objectives directly and to provide a controlled, privacy-preserving foundation for evaluating FL performance in fraud detection.~\cite{jpmorgan2025aikya}.

\paragraph{Rationale for Synthetic Data.}

The fundamental obstacle in developing robust payment anomaly detection models -- particularly in federated settings -- is the scarcity of labeled, high-fidelity transaction data. Real-world payment datasets are subject to stringent regulatory and privacy constraints that preclude open sharing, and even when accessible, exhibit extreme class imbalance: fraudulent transactions typically constitute less than 1\% of total volume. Federated learning compounds this difficulty by requiring heterogeneous data distributions across participants to validate the premise that collaborative training over diverse local distributions yields superior global models. To address these constraints, we develop a fully configurable, purely synthetic data generation framework. The framework is designed with the following properties:

\begin{itemize}
    \item Controllable statistical heterogeneity: Each simulated institution (site) draws feature values from independently parameterized distributions, enabling systematic variation in data characteristics across federated participants.
    \item Composable, rule-based anomaly injection: Anomalies are introduced through a modular set of perturbation functions that can be applied independently or composed to create multi-dimensional fraud signatures of varying complexity.
    \item Adjustable class balance and label noise: The fraction of anomalous records and the probability of correct labeling are both configurable, allowing the generation of datasets spanning the spectrum from idealized (high anomaly fraction, perfect labels) to realistic (extreme imbalance, noisy labels).
\end{itemize}

\paragraph{Cross-silo non-IID Distribution Choices and Parameter Configuration}
To emulate heterogeneity across financial institutions, each client site is parameterized by (i) a subset of anomaly types observed during training and (ii) site-specific feature distributions. The generator configuration defines multiple sites with differing anomaly-type exposure, such that no single site observes all anomaly types. This induces a non-IID federated learning setting in which cross-site collaboration can improve detection of underrepresented or unseen anomaly types.

\paragraph{Feature distributions and tunable parameters.}
The generator uses simple, interpretable distributions that are independently configurable per site:

\begin{itemize}
    \item Transaction amounts use the \emph{log-normal distribution} exclusively, for three reasons: (i) it guarantees strictly positive, finite samples -- a normal distribution $\mathcal{N}(\mu, \sigma^2)$ assigns $P(X < 0) = \Phi(-\mu/\sigma)$, yielding negative amounts for non-trivial $\sigma/\mu$; (ii) its right-skewed, heavy-tailed shape matches empirical payment amount profiles, unlike uniform distribution which would produce unrealistic flat densities; and (iii) it arises naturally from multiplicative processes (exchange rates, quantity-price products) via the multiplicative CLT. The Gamma distribution shares non-negativity but its shape/scale parameterization is less intuitive than specifying arithmetic means directly, and its lighter tail under-covers extreme anomalous amounts.
    \item Coordinate perturbations and categorical selections use \emph{uniform distributions}, encoding maximum ignorance within bounded intervals -- no tower offset magnitude is preferred over another, and signed perturbations preclude log-normal. \item Discrete uniform sampling for activity event counts forces the classifier to learn range boundaries rather than exploit distributional shape.
\end{itemize}
Log-normal distributions are specified as desired arithmetic mean $\bar{x}$ and standard deviation $\sigma_{\text{raw}}$, converted to underlying parameters via:
$$\sigma^2 = \ln\!\left(1 + \frac{\sigma_{\text{raw}}^2}{\bar{x}^2}\right), \qquad \mu = \ln(\bar{x}) - \frac{\sigma^2}{2}$$
Cross-site heterogeneity is then achieved by varying these parameters (e.g., normal personal amounts: $\bar{x} = 20{,}000$ at \texttt{site-A} vs. $\bar{x} = 150{,}000$ at \texttt{site-E}) and tower perturbation bounds ($[-0.75, 1.25]^\circ$ vs. $[-10, 10]^\circ$).

\paragraph{Base data generation}

Each record comprises several attributes

\begin{itemize}
    \item Participant attributes (debitor/creditor): Basic identity (names, DOB, address), physical geo-coordinates from real land locations are chosen via Faker.
    \item Account creation timestamp uniformly in $(t_{\text{dob}}, t_{\text{now}} - 12\text{w}]$, last activity in $[\max(t_{\text{create}}, t_{\text{now}} - 1\text{w}), t_{\text{now}}]$
    \item Account type is chosen uniformly over \texttt{\{SAVINGS, CHECKING, BUSINESS\}}
    \item Tower coordinates perturbed from physical location: $lat_{\text{tower}} = lat_{\text{phys}} + \Delta$, $\Delta \sim \text{Uniform}(l, h)$.
    \item Activity events are drawn from account-type-conditioned ranges (\texttt{BUSINESS}: $[50, 1.5\text{M})$, \texttt{CHECKING}: $[2, 500)$, \texttt{SAVINGS}: $[2, 50)$)
    \item Financial attributes: Debitor amount from site-specific log-normal conditioned on account type, Creditor amount via $A_{\text{crdtr}} = A_{\text{dbtr}} \cdot r(C_{\text{dbtr}}, C_{\text{crdtr}})$ using ECB exchange rates.
    \item All rows initialize with \texttt{FRAUD\_FLAG = 0}.
\end{itemize}

\paragraph{Transaction amount distributions.}
Transaction amounts are modeled with log-normal distributions to reflect the heavy-tailed nature of real-world payment values. 
Faker is used for identities/addresses; temporal fields follow the parameterized sampling rules described above; amounts are generated via site-specific log-normal distributions.
    
\paragraph{Anomaly-type Rules.}
We focus on four structured anomaly families used throughout the experiments. Attributes are grouped based on anomaly types

\begin{itemize}
    \item \emph{Tower/Physical Location Mismatch (Type 1)} - Tower coordinates are displaced beyond the normal range of values. For coordinate $c \in [c_{\min}, c_{\max}]$, available displacement $\delta = (c_{\max} - c) + p$ where $p$ is a site-specific (typically negative) perturbation factor that pushes the new coordinate toward range extremes. Candidate ranges $R = (c_{\max} - \delta, c_{\max})$ or $(c_{\min}, \delta - |c_{\min}|)$ are sampled uniformly. Applied to all four tower coordinates (lat/lon for both participants).
    \item \emph{Type 2 - Immature Account + Anomalous Amount} - Here account creation is reset to minutes before payment: $t_{\text{create}}^{\text{new}} = t_{\text{init}} - (h \cdot 3600 + m \cdot 60 + s)$ where $h \in \{0..5\}$, $m \in \{0, 1, 4, 9, 16, 25\}$, $s \in \{1, 10, 20, 30, 40, 50\}$. Amount is resampled from the anomalous log-normal (e.g., \texttt{site-E}: $\bar{x}_{\text{anom}} = 750{,}000$ vs. $\bar{x}_{\text{normal}} = 150{,}000$).
    \item \emph{Type 3 - Dormant Account Reactivation} - Last activity is pushed $d \in \{90, 120, 150, 180\}$ days before payment (plus sub-day jitter), creating 3--6 month dormancy gaps.
    \item \emph{Type 4 - Anomalous Transaction Frequency} - Activity counts elevated to: \texttt{BUSINESS} $[1\text{M}, 5\text{M})$, \texttt{CHECKING} $[525, 2000)$, \texttt{SAVINGS} $[75, 2000)$ -- overlapping minimally with normal ranges.
\end{itemize}

\begin{table}[t]
\centering
\caption{\textbf{Anomaly types used across client sites.} Structured anomaly families injected into the synthetic payment dataset.}
\label{tab:anomaly_types}
\small
\setlength{\tabcolsep}{5pt}
\renewcommand{\arraystretch}{1.15}
\begin{tabularx}{\textwidth}{@{}l X X@{}}
\toprule
\textbf{Anomaly type} & \textbf{Description} & \textbf{Example} \\
\midrule
\textbf{1:} Location-based &
Significant geographical deviation between expected and actual transaction location &
Expected location: 500 miles from NYC; origin: Azerbaijan \\
\textbf{2:} Account-age-based &
High-value transactions from disproportionately new accounts &
Account age: 5 hours; amount: \$15{,}000 \\
\textbf{3:} Inactivity-based &
Unusual transactions following extended periods of account inactivity &
Last activity: 6 months ago; sudden high/low-value transaction \\
\textbf{4:} Unusual activity for account type &
Excessive transaction frequency over a short time window &
50 transactions in 10 minutes for a \texttt{CHECKING} or \texttt{SAVINGS} account \\
\bottomrule
\end{tabularx}
\end{table}

\paragraph{Row selection and Label Noise}

Row selection supports controlled overlap parameterized by percentage of fraud rows $\alpha$ and fraud type overlap fraction $\beta$. If $\mathcal{D}$ is the dataset, with $\beta > 0$, $n_F = \lceil \lceil |\mathcal{D}| \alpha \rceil \beta \rceil$ rows are drawn from previously-flagged records and $n_N = \lceil |\mathcal{D}| \alpha \rceil - n_F$ from clean records, enabling composite multi-type anomalies. Post-injection, label noise is applied: assuming $p_\text{apply}$ is the fraction of rows where fraud labels should be applied, a fraction $(1 - p_{\text{apply}})$ of fraud-flagged randomly sampled rows have labels flipped to $0$, producing records with anomalous features but non-fraudulent labels.

The final dataset generated with $n$ rows is highly imbalanced with total fraudulent rows $n_F$, $\frac{n_F}{n} \in [0.001, 0.01)$.

\paragraph{Client-side Pre-processing.}

From $\approx50$ raw attributes, a feature vector is engineered:
\begin{itemize}
    \item Temporal features: These are derived as account age in minutes ($\frac{t_{\text{init}} - t_{\text{create}}}{60}$) for debitor and creditor, and last-activity gap in seconds, then log-transformed ($x' = \ln x$) to compress the multi-order-of-magnitude range between normal ($\sim 10^5$ min) and anomalous ($\sim 1$ min) account ages.
    \item Geospatial features: We compute Haversine distance between physical and tower coordinates
    $$a = \sin^2\!\left(\frac{\Delta lat}{2}\right) + \cos(lat_1)\cos(lat_2)\sin^2\!\left(\frac{\Delta lon}{2}\right), \quad d = 2R \cdot \text{atan2}(\sqrt{a}, \sqrt{1-a})$$
    with $R = 6334.08$ km.
    
    \item Categorical encoding: account type is one-hot encoded into three binary indicators (\texttt{BUSINESS}, \texttt{CHECKING}, \texttt{SAVINGS}).
    
    \item Scaling: The numerical features are scaled via \texttt{StandardScaler}
    \item Target: \texttt{FRAUD\_FLAG} $\in \{0,1\}$
\end{itemize}



\begin{table}[t]
\centering
\caption{\textbf{Features of interest and associated anomaly type.} Only a subset of the 60+ features is shown (most relevant for the anomaly rules).}
\label{tab:feature_types}
\small
\setlength{\tabcolsep}{5pt}
\renewcommand{\arraystretch}{1.15}
\begin{tabularx}{\textwidth}{@{}l l X l@{}}
\toprule
\textbf{Feature} & \textbf{Type} & \textbf{Description} & \textbf{Data type} \\
\midrule
FRAUD\_FLAG & Target & Indicates whether the transaction is anomalous (historically labeled \texttt{FRAUD}). & int \\
\midrule
DEBITOR\_GEO\_LATITUDE  & Type 1 & Latitude of debitor's expected location. & float \\
DEBITOR\_GEO\_LONGITUDE & Type 1 & Longitude of debitor's expected location. & float \\
DEBITOR\_TOWER\_LATITUDE  & Type 1 & Latitude of cell tower used in transaction. & float \\
DEBITOR\_TOWER\_LONGITUDE & Type 1 & Longitude of cell tower used in transaction. & float \\
\midrule
PAYMENT\_INIT\_TIMESTAMP & Type 2 & Timestamp when payment was initiated. & datetime \\
DEBITOR\_AMOUNT          & Type 2 & Amount debited from debitor's account. & float \\
DEBITOR\_ACCOUNT\_CREATE\_TIMESTAMP & Type 2 & Account creation timestamp for debitor. & datetime \\
\midrule
DEBITOR\_ACCOUNT\_LAST\_ACTIVITY\_TIMESTAMP & Type 3 & Last activity timestamp for debitor's account. & datetime \\
\midrule
DEBITOR\_ACCOUNT\_TYPE & Type 4 & Type of debitor's account (e.g., checking, savings). & str \\
DEBITOR\_ACCOUNT\_ACTIVITY\_EVENTS\_PAST\_30D & Type 4 & Number of account activity events in the past 30 days. & int \\
\bottomrule
\end{tabularx}
\end{table}

\subsection{Model and Training}

\paragraph{Deep Learning versus Traditional Machine Learning:}

Deep neural networks (DNNs) were selected over conventional machine learning models (e.g., logistic regression, random forests, XGBoost) for several methodological and practical considerations. First, DNNs offer superior representational capacity, enabling the automatic extraction of complex, nonlinear feature interactions from high-dimensional transactional data without extensive manual feature engineering. Second, their gradient-based optimization and continuous parameter spaces align naturally with federated aggregation strategies such as FedAvg, rendering DNNs particularly compatible with FL paradigms.

Moreover, the hierarchical feature representations learned by DNNs support more effective knowledge transfer across heterogeneous, non-IID client datasets—a common characteristic in multi-institutional financial settings. From an infrastructure standpoint, DNNs scale efficiently on modern accelerators, including the distributed GPU resources available through NVIDIA DGX Cloud. Finally, deep learning methods consistently achieve state-of-the-art performance in fraud detection, reinforcing their suitability for deployment in production financial systems.

\paragraph{Network Architecture Choice:}

The chosen architecture, comprising three fully connected hidden layers with 64, 32, and 16 neurons respectively, reflects a deliberate balance between model expressiveness and operational constraints. This configuration provides sufficient capacity to model diverse fraud patterns while maintaining a compact parameterization that mitigates overfitting--an important consideration under non-IID data distributions. This small model footprint also reduces communication overhead during federated parameter exchanges, thereby improving round-trip efficiency. Additionally, networks of this scale exhibit favorable convergence behavior, decreasing the number of federated training rounds required to achieve stable performance.

\paragraph{Model Summary:}
\begin{itemize}
\small
    \item \textbf{Model Type:} Deep Neural Network (DNN) implemented in PyTorch.
    \item \textbf{Architecture:} Fully connected feed-forward network consisting of:
    \begin{itemize}
        \item \textbf{Input Layer} (Size: Number of input features)
        \item \textbf{Hidden Layer 1:} 64 neurons (LayerNorm + ReLU activation)
        \item \textbf{Hidden Layer 2:} 32 neurons (LayerNorm + ReLU activation)
        \item \textbf{Hidden Layer 3:} 16 neurons (LayerNorm + ReLU activation)
        \item \textbf{Output Layer:} 2 neurons
    \end{itemize}
    \item \textbf{Hyperparameters:} Learning Rate = 5e-4, FL Rounds = 20, Epochs per Round = 1, Batch Size = 64, Loss Function = FocalLoss.
    \item \textbf{Workflow:} The experiment followed the NVFlare Scatter-Gather workflow, where the server initialized the global model from scratch and aggregated model updates (e.g., via FedAvg), and clients performed local training on their private datasets.
\end{itemize}

\subsection{Sample-level Differential Privacy}

We enforced sample-level $(\varepsilon,\delta)$–differential privacy during local training using the \texttt{PrivacyEngine} module from Opacus. Differential privacy requires that, for any two datasets $\mathcal{D}$ and $\mathcal{D}'$ differing in a single sample, and for any measurable set of outputs $\mathcal{O}$,

$$\Pr[\mathcal{A}(\mathcal{D}) \in \mathcal{O}] 
\le e^{\varepsilon}\, \Pr[\mathcal{A}(\mathcal{D}') \in \mathcal{O}] + \delta,$$

where $\mathcal{A}$ denotes the randomized learning algorithm.

To satisfy this guarantee, we implement the Gaussian Mechanism with per-sample gradient clipping. For a model with parameters $\theta$, we modify each training step as follows:

\paragraph{1. Per-sample gradients.}
Let $g_i$ denote the gradient contributed by sample $i$ in a minibatch $B$. The gradients are clipped to a fixed $\ell_2$ norm bound $C$:

$$\tilde{g}_i = g_i \cdot \min\left(1, \frac{C}{\lVert g_i \rVert_2}\right).$$

\paragraph{2. Addition of calibrated Gaussian noise.}
The aggregated gradient for the minibatch is then perturbed as

$$\hat{g} = \frac{1}{|B|} \sum_{i \in B} \tilde{g}_i + \mathcal{N}(0, \sigma^2 C^2 I),$$

where $\sigma$ is the noise multiplier.

\paragraph{3. Privacy accounting.}
Opacus uses Rényi Differential Privacy (RDP) composition to track the cumulative privacy loss across all optimization steps. Given a target $(\varepsilon, \delta)$ and a total number of epochs $E$, the privacy engine automatically selects a noise multiplier $\sigma$ such that the final RDP guarantee converts to the desired $(\varepsilon, \delta)$ bound.

\paragraph{Integration into training.}
In practice, the training loop is privatized with:
\begin{verbatim}
from opacus import PrivacyEngine
...
model, optimizer, train_loader = privacy_engine.make_private_with_epsilon(
    module=model,
    optimizer=optimizer,
    data_loader=train_loader,
    epochs=E,
    target_epsilon=1.0,
    target_delta=1e-5,
    max_grad_norm=1.0,
)
\end{verbatim}
which wraps the model, optimizer, and data loader with the clipping, noise injection, and privacy accounting described above.

This procedure ensures that each institution's contribution remains differentially private at the level of individual training samples, providing strong formal privacy guarantees while preserving the overall learning dynamics.

\subsection{Evaluation}
We evaluated model performance using F1-score as the primary metric, given its ability to balance precision and recall in fraud-detection settings where both false positives and false negatives carry significant cost. For per-type analysis, we treat each anomaly type as a separate \emph{binary} detection task (type-$k$ vs.\ non-fraud) and compute binary F1 with fraud as the positive class ($y{=}1$). Predicted labels are obtained via \texttt{argmax} over the model outputs, and we report the mean F1 aggregated across anomaly types and client sites. For a more complete assessment, we also compute Accuracy, Precision, and Recall as secondary metrics. Model training progress was monitored through convergence behavior, defined as the point at which performance metrics stabilized across successive communication rounds within a predefined variance threshold. After convergence, we assessed model interpretability using Shapley-value–based attribution (Captum GradientShap) to verify that the learned feature attributions aligned with known patterns of anomalous financial activity (using 500 samples for baseline attribution).

\paragraph{Evaluation Summary:}
\begin{itemize}
\small
    \item \textbf{Primary Metric:} Binary F1-score for fraud detection, with fraud as the positive class ($y{=}1$). Predicted class labels are obtained via \texttt{argmax} over the model outputs. Secondary metrics include Accuracy, Precision, and Recall.
    \item \textbf{Convergence Criteria:} Performance metrics stabilize across consecutive rounds ($\sigma \le$ threshold).
    \item \textbf{Interpretability:} Model decisions were analyzed using Shapley-value--based attribution post-convergence to confirm feature importance alignment with expected anomaly rules.
\end{itemize}

\paragraph{Controlled Comparison:}
To ensure a fair comparison between Local, Centralized, and Federated (FedAvg) training, we keep the experimental protocol identical across all settings. All models use the same network architecture and the same optimizer/hyperparameters (e.g., learning rate, batch size, and number of local epochs). 
Each client applies the same feature preprocessing pipeline (including normalization/scaling) using parameters fit on its scaling dataset (derived from training data only; no test leakage), and the same set of input features is used in every run. 
Evaluation is performed on a held-out test split using the same metric implementation procedure across Local, Centralized, and Federated models, so observed differences in performance can be attributed to the training paradigm rather than changes in data processing, model capacity, or evaluation methodology.

\subsection{Shapley Value–based Feature Attribution}

Shapley values provide a principled mechanism for attributing a model prediction to individual input features. For a model $f$ with feature set $F$, the Shapley value $\phi_i$ for feature $i$ is defined as

$$\phi_i = \sum_{S \subseteq F \setminus \{i\}}
\frac{|S|!\, (|F|-|S|-1)!}{|F|!}
\left[ f_{S \cup \{i\}}(x) - f_{S}(x) \right].$$

This expression computes the marginal contribution of feature $i$ to every possible coalition $S$, weighted so that all feature orderings are treated symmetrically.

For a given input $x$, we express the model output as

$$f(x) = \phi_0 + \sum_{i=1}^{M} \phi_i,$$

where $\phi_0$ is the expected model prediction and $M$ is the number of features. This additive decomposition satisfies efficiency, symmetry, dummy, and additivity—the axioms uniquely characterizing Shapley values.

Exact computation is exponential in $|F|$, so practical implementations rely on approximations. In this work, we use \texttt{GradientShap} from Captum\footnote{\url{https://captum.ai/}}~\cite{kokhlikyan2020captum}, which estimates attributions for differentiable models by integrating input gradients along the path between a baseline $\mathbf{x}_{\mathrm{base}}$ and the input $\mathbf{x}$:

$$\phi_i \approx \frac{1}{K} \sum_{k=1}^{K}
(x_i - x_{i,\mathrm{base}})
\left. \frac{\partial f}{\partial x_i} \right|_{\mathbf{x}^{(k)}},$$

where $\mathbf{x}^{(k)}$ are interpolated samples. This yields faithful, locally accurate feature attributions suitable for regulated financial applications.

\subsection{Federated Learning Software}

Next, we outline how NVFlare was configured, distributed, and executed across participating financial institutions, including secure provisioning, job orchestration, dependency management, and real-time monitoring. Together, these components form a reproducible, production-ready software environment that enables reliable experimentation at scale.

\paragraph{Software Distribution and System Operation:} The FL framework was deployed and managed using the NVFlare Dashboard, which provides a centralized interface for software distribution, configuration, and orchestration across all participating sites. This approach ensures consistency, security, and reproducibility in multi-institution federated experiments.

\paragraph{Startup Kit Generation:}
Administrators used the NVFlare Dashboard to generate cryptographically signed startup kits for the server and each client. These kits include configuration files, authentication credentials, and certificates required for secure onboarding.

\paragraph{Client Provisioning:}
Startup kits were securely transmitted to each participating institution, where clients installed the NVFlare package and initialized their environment. The configuration supplied within each kit specifies network endpoints, resource allocation, and institution-specific security policies.

\paragraph{Job Submission:}
Experimental workflows—comprising model architectures, hyperparameters, training scripts, and data interfaces—were packaged as NVFlare jobs. Jobs were submitted through the Dashboard, which automatically distributed them to all clients. Integrated version control and rollback mechanisms ensured safe and reproducible deployment of updates.

\paragraph{Package and Dependency Management:}
Python dependencies (e.g., PyTorch, Captum, Pandas) were managed via dedicated Docker containers. Custom training code and model definitions were encapsulated as Python modules, and NVFlare automatically synchronized these components across all sites. The distributed package set included client and server executables, model definitions (e.g., \texttt{model.py}), training logic (e.g., \texttt{client.py}, \texttt{central\_train.py}), data preprocessing utilities, and evaluation or visualization scripts.

This Dashboard-based distribution mechanism standardizes execution across institutions and reduces the operational complexity inherent to multi-party FL deployments.

\paragraph{Real-Time Monitoring Framework:}
To enable comprehensive observability throughout the federated training process, we integrated MLFlow as a real-time experiment tracking and visualization system.
The MLFlow interface provided real-time visualization of global and per-site performance, enabling comparison among different federated optimization algorithms (e.g., FedAvg, FedProx, FedOpt). The dashboard facilitated early detection of convergence issues, client dropouts, or data anomalies and supported systematic hyperparameter optimization.
This monitoring framework offers essential transparency into the behavior of the federated system, supporting both scientific rigor and operational reliability.

\paragraph{Server Deployment:}
An MLFlow tracking server was provisioned on NVIDIA DGX Cloud with persistent storage and secure web-based access for authorized collaborators. All metrics, hyperparameters, and artifacts were centrally recorded to ensure reproducibility and auditability.

\paragraph{Client Integration:}
Each NVFlare client logged training metrics to the MLFlow server during federated rounds. Automatically logged outputs included training and validation losses, class-specific performance metrics (precision, recall, F1-score), confusion matrices, Shapley-value–based attribution (GradientShap) interpretability plots, hyperparameters, model checkpoints, and measurements of runtime and communication overhead.

\paragraph{System Architecture:}
The experimental system employs a client–server federated architecture implemented using NVFlare. Each client—representing a financial institution—trains a local model on proprietary datasets that remain on-premise. A central NVFlare server aggregates local model updates to form a global model, enabling collaborative learning without sharing raw data.

\subsection{Federated System Infrastructure}
The experiment utilized a multi-site federated topology on NVIDIA DGX Cloud\footnote{\url{https://www.nvidia.com/en-us/data-center/dgx-cloud} \label{fn:dgx}} infrastructure as shown in Table~\ref{tab:dgx}. The NVIDIA DGX Cloud infrastructure offers several advantages that were critical to the execution of this study. First, its fully managed GPU clusters substantially reduce operational burden, enabling researchers to focus on methodological development rather than system administration. Second, the platform incorporates enterprise-level security and compliance mechanisms, making it suitable for research involving sensitive financial applications. Third, DGX Cloud supports dynamic, on-demand scaling of compute resources, facilitating experiments of varying size and complexity. Fourth, the environment is built on the latest generation of NVIDIA GPU technology, providing the high-performance acceleration needed for efficient deep neural network training. Finally, the platform allows easy integration with NVIDIA FLARE, including optimized networking configurations that streamline FL deployments.

To enable all participants to run on the NVIDIA DGX Cloud, a Docker image was built that includes all software dependencies, such as Python packages, drivers, libraries, and configurations. The PyTorch container image from NVIDIA NGC Catalog was used as the base image. We summarize the system setup in Table~\ref{tab:dgx}.

\begin{table}[t]
\centering
\caption{System setup and infrastructure on NVIDIA DGX Cloud.}
\label{tab:dgx}
\small
\setlength{\tabcolsep}{5pt}
\renewcommand{\arraystretch}{1.0}
\begin{adjustbox}{max width=0.8\linewidth,center}
\begin{tabularx}{\linewidth}{@{}l X@{}}
\toprule
\textbf{Component} & \textbf{Specification} \\
\midrule
Deployment model & Multi-site federated topology on DGX Cloud \\
Server & One NVFlare server instance (CPU-only) \\
Clients & Five client sites, each with a dedicated NVIDIA H200 GPU instance (DGX Cloud managed) \\
FL framework & NVIDIA FLARE (NVFlare) SDK \\
Runtime & Docker image based on an NVIDIA NGC PyTorch container \\
\bottomrule
\end{tabularx}
\end{adjustbox}
\end{table}

\section*{Code \& Data Availability}
Code for data generation and FL with NVIDIA FLARE is available\footnote{\url{https://nvidia.github.io/NVFlare/research/fsi-fraud-detection}}.

\section*{Acknowledgments}
We gratefully acknowledge NVIDIA's support through DGX Cloud\footref{fn:dgx}, which enabled this research.

\bibliography{references}

\end{document}